\documentclass[letterpaper, 10 pt, conference]{ieeeconf}
\IEEEoverridecommandlockouts                              %
\usepackage{graphicx} %
\usepackage{epsfig} %
\usepackage{mathptmx} %
\usepackage{amsmath} %
\usepackage{amssymb}  %
\usepackage{xcolor}
\usepackage[hidelinks]{hyperref}
\usepackage{url}
\usepackage{soul}

\usepackage{mathrsfs}
\usepackage{subcaption}
\usepackage{float}
\usepackage[11pt]{moresize}
\usepackage[english]{babel}
\usepackage[utf8]{inputenc}
\usepackage{csquotes}
\usepackage{makecell}

\usepackage{colortbl}

\title{
On the Visual-based Safe Landing of UAVs in Populated Areas: a Crucial Aspect for Urban Deployment.
}

\begin{document}

\author{Javier González-Trejo, Diego Mercado-Ravell*, Israel Becerra and Rafael Murrieta-Cid%
\thanks{This work was supported by the Mexican National Council of Science and Technology CONACyT, and the FORDECyT project 296737 “Consorcio en Inteligencia Artificial”. }%
\thanks{Authors are with the Center for Research in Mathematics CIMAT AC, Mexico. D. Mercado-Ravell \& I. Becerra are also with Cátedras CONACyT (*corresponding author email: {\tt\footnotesize diego.mercado@cimat.mx}).}}
\maketitle
\begin{abstract}
Autonomous landing of Unmanned Aerial Vehicles (UAVs) in crowded scenarios is crucial for successful deployment of UAVs in populated areas, particularly in emergency landing situations where the highest priority is to avoid hurting people. In this work, a new visual-based algorithm for identifying Safe Landing Zones (SLZ) in crowded scenarios is proposed, considering a camera mounted on an UAV, where the people in the scene move with unknown dynamics. To do so, a density map is generated for each image frame using a Deep Neural Network, from where a binary occupancy map is obtained aiming to overestimate the people's location for security reasons. Then, the occupancy map is projected to the head's plane, and the SLZ candidates are obtained as circular regions in the head's plane with a minimum security radius. Finally, to keep track of the SLZ candidates, a multiple instance tracking algorithm is implemented using Kalman Filters along with the Hungarian algorithm for data association. Several scenarios were studied to prove the validity of the proposed strategy, including public datasets and real uncontrolled scenarios with people moving in public squares, taken from an UAV in flight. The study showed promising results in the search of preventing the UAV from hurting people during emergency landing.
\end{abstract}


\IEEEpeerreviewmaketitle

\section{Introduction}

Over the last decades, the world have witnessed a great interest in the development and production of Unmanned Aerial Vehicles (UAVs), flooding the market with all kinds of drones for different budgets and with continuously increasing capabilities. As a direct consequence, the number of successful civilian applications has been steadily increasing, and everyday, it is more common to see UAVs integrated in our daily lives, interacting closer to the people in a wide variety of tasks ranging from surveillance, photography, structure inspection, load transportation, fast response in disasters, to simple entertainment, among others \cite{nex13_uav_mappin_applic} \cite{lucieer13_mappin_lands_displ_using_struc} \cite{menouar17_uav_enabl_intel_trans_system_smart_city}.

Nevertheless, the increased use of UAVs has also raised several well founded legal and public security concerns, in part due to the threat of violating privacy, but mainly derived from the inherent risk of an accident produced by a system failure or human error. Although big efforts have been devoted to increase the robustness of these systems, e.g., safety protocols for returning home have helped to attenuate security issues, it is not enough to warrant people's safety on the streets in case of a system failure. Consequently, this has restricted the use of drones to rural environments, or wide urban areas clear of people, precluding some very interesting applications in populated zones such as surveillance of crowds, monitoring mass events, fast response in health emergencies, criminals chasing and law enforcing, etc. \cite{motlagh17_uav_based_iot_platf}.

Accordingly, the legislation generally prohibits any UAV activity over a crowd \cite{lawDrones}, but in practical terms, it is unavoidable that an UAV will flight above an unaware crowd at lower altitudes putting at risk the persons safety in case of any UAV malfunctioning, such a communication loss or power shortage, or even a human error. Henceforth, it is of crucial relevance to further provide every UAV with safety protocols to at least avoid hurting people in case of an emergency, being able to autonomously find landing zones without putting people at risk. If every UAV were provided with such an emergency autonomous landing system, this would not only help to prevent hurting, or even killing people during drones' crashes, but it also would considerably boost all their potential for successful urban deployment, even in crowded scenarios.

\begin{figure}[t]
\centerline{\includegraphics[width=0.45\textwidth]{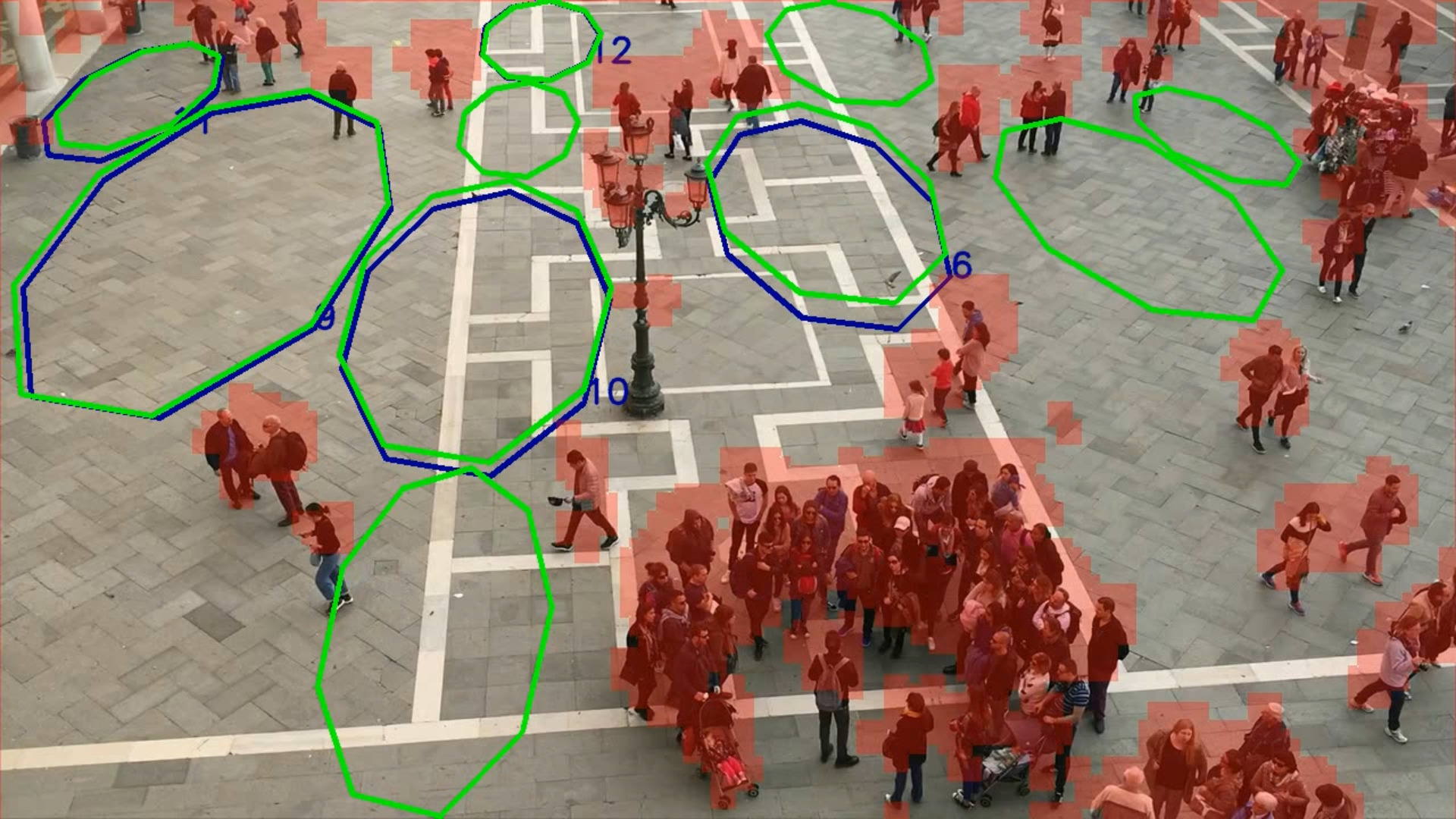}}
\caption{Autonomous visual-based Safe Landing Zone (SLZ) proposals in populated areas. In the case of a system failure, UAVs must at least avoid hurting people. The red regions are places where people is detected using density maps. The green circles represent the regions free of people where the drone can land. Finally, the blue circles indicate landing zones tracked by KFs to ensure time consistency.}
\label{fig:introduction}
\end{figure}

The idea of visual-based autonomous landing of flying machines has been revisited in the past, and some examples are to be found mainly to locate predefined landing tags \cite{saripalliil_vision} \cite{garcia2002towards}, or to identify and avoid hazardous terrain \cite{johnsonil_vision_guided_landin_auton_helic_hazar_terrain}. Nevertheless, autonomous safe landing in crowded scenarios is an almost unexplored and very challenging task, due to the implied security constrains  and the inherent difficulty of the problem, where people may cluster in dense groups producing severe occlusions, while freely moving in the scene with unknown dynamics. It is only thanks to recent advancements in modern deep learning techniques for computer vision that we can attempt to solve this complicated task, where just a few related works are to be found in the literature. 

In \cite{tzelepi19_graph_embed_convol_neural_networ}, the authors propose a new neural network and regularization technique based on the graph embedding network to detect what they call no-fly zones, or zones where a crowd is present on the scenario. Moreover, in \cite{liu19_geomet_physic_const_drone_based} a new neural network is proposed for time consistency across the frames in a video, taking into account the perspective distortion caused by the projection transformation from the real world to the image. The Safe Landing Zones (SLZ) are a byproduct of the density consistency achieved on the work. More recently, in \cite{castellano20_crowd_detec_aerial_images_using}, a model with a MobileNet back-end is presented, specifically tailored for images coming from an UAV, and fast enough for real time applications. However, all these works are limited to segment the scene in areas free of people, although this is a necessary and important first step for the autonomous safe landing in crowded areas, none of them attempts to solve the problem of indicating which regions are the more appropriate to safely land the vehicle.
It is important to stress that we do not use classical methods to avoid hazardous areas, because in this work the priority is to avoid landing over people and not to ensure the drone's integrity.
Furthermore, an important aspect that is not considered in works like the one in \cite{johnsonil_vision_guided_landin_auton_helic_hazar_terrain} is that people move.

In that regard, this paper presents an algorithm to detect Safe Landing Zones (SLZ) in real populated and uncontrolled scenarios, considering a camera mounted on an UAV, and where people are moving with unknown dynamics, prioritizing their integrity in emergency situations, as depicted in Fig. \ref{fig:introduction}. To do so, we consider the use of a Deep Neural Network (DNN) density map generator to infer people's location on the image, then, an occupancy binary mask is obtained aiming at overestimating the people's location for security reasons. In our previous work in \cite{gonzalezLandingSite}, we proposed and implemented on a mobile phone a lightweight architecture for real-time density map generation, from where, as a first approach, the biggest circle free of people was found directly on the image plane using the Polylabel algorithm, without considering the camera movement. In contrast, in this work, the occupancy map is projected to the head's plane in the real world, where multiple SLZ candidates are obtained as circular regions with a radius larger than a minimum value $r_0$. 

Furthermore, by means of Kalman Filters (KF) and the Hungarian algorithm for data association \cite{bruff2005assignment}, a multiple instances tracking technique  is implemented using the SLZ candidates as measurements, smoothing the SLZ proposals movement and providing time consistency along frames. Between two times instants $k$ and $k+1$, a KF for each landing zone gives an estimate of the location and radius of a landing zone in the head's plane. The Hungarian algorithm finds the most similar landing zone a time $k+1$ for a given zone at time $k$; that landing zone will be the one that maximizes the area of intersection over the union area (IoU) of the landing zones a time $k$ and $k+1$. Indeed, the Hungarian algorithm orders the similarity of all possible matching using that criterion.

Lastly, several scenarios were studied from available crowd datasets, as well as in real uncontrolled and challenging scenarios recorded from a flying drone in public squares with people in continuous movement. Although further work is still needed to obtain a fully reliable solution, the proposed strategy showed good performance in spite of the challenging scenarios, proving to be helpful to prevent hurting people in case of emergency landing.

The reminder of the paper is structured as follows. In Section \ref{sec:crowds}, we briefly introduce the density map generators and their utility for detecting crowds in images. Section \ref{sec:proposals} describes the SLZ proposal generator in the head's plane. Section \ref{sec:tracking} presents the multiple instances tracking algorithm. In Section \ref{sec:experiments}, the experiments' results are presented. Finally, in Section \ref{sec:conclusions}, conclusions and future work are discussed.

\section{Crowds Detection using Density Maps}
\label{sec:crowds}
In order to find suitable landing zones where no person could be harmed, it is necessary to distinguish the crowd from a safe landing zone. One of the most popular methods for crowd detection are the density map generators \cite{kang2018beyond}. Density maps are obtained by means of DNN trained over crowd images containing head annotations, and provides spatial information on the crowd location even subject to large occlusions, furthermore, it provides an estimated of the number of persons in an image or scene as the integral of the corresponding density map \cite{zhang16_singl_image_crowd_count_multi}. These density maps have excellent results in the counting task for a wide variety of crowded scenes, ranging from sparse to highly dense crowds, where person-to-person occlusion could be so severe to the point that the only visible part of a person is a portion of the head. 

In contrast to the classical detect-and-count approaches \cite{bochkovskiy2020yolov4}, \cite{tan2019efficientdet}, which try to detect the whole body for each person, the density map generators are trained to detect the heads of persons in a crowd \cite{zhang16_singl_image_crowd_count_multi}, increasing the robustness against occlusion, at the cost of being more prone to overestimate the people's locations under rich textured backgrounds.
The first works using density maps with deep learning algorithms are found in \cite{zhang16_singl_image_crowd_count_multi}, with the algorithm Multi Column Convolutional Neural Network (MCNN) that used multiple columns to capture different perspectives and  heads' scales. The latest works focus not only on obtaining more accurate architectures, but also on improving the manual annotations through learning \cite{bai2020adaptive}, or creating new loss functions to better learn how to count and locate the crowd in an image \cite{wang20_distr_match_crowd_count}.

The accuracy of people's location is of great relevance for the SLZ detection task in order to avoid human accidents. In that sense, due to the inherent security constrains, it is preferred to overestimate the people's location, hence, it is preferred to have false positive detections, but it is unacceptable to miss detection that can result in hurting a person. Henceforth, the employed density map generator should be explicitly trained to overestimate the people's locations to increment the security. 

Another important aspect to take into account when selecting the density map generator is its  computational complexity and time response. It is of crucial importance to have the information at a high rate in order to perform the landing maneuver in real-time, preferably embedded on the vehicle. Current state-of-the-art algorithms, especially the works using pre-trained DNN back-ends like VGG (Visual Geometry Group), are slow, mostly in embedded devices, which hinders the objective of implementing them in most UAVs. Accordingly, it is required to have a good trade-off between accuracy and computational cost when selecting the density map generator, where recent advancements in lightweight architectures appear as a good alternative for the SLZ detection task. These lightweight architectures could be defined as DNN below the 1 million of parameters, being suitable to be implemented in real-time solutions for both high-end computers and embedded devices \cite{gonzalezLandingSite}. For this work, we have selected our own custom lightweight density map generator called Pruned BL CCNN \cite{gonzalezLandingSite}, but any other state-of-the art density map generator can be used instead depending on the available hardware. The Pruned BL CCNN obtained good results in the crowd counting and detection tasks, while being light enough to be implemented for real-time application on a mobile phone app. This network was explicitly trained to overestimate the crowd for this particular task, in order to increase the safety of our algorithm.

\section{Safe Landing Zone Proposals}
\label{sec:proposals}

\begin{figure}[t]
\centerline{\includegraphics[width=0.5\textwidth]{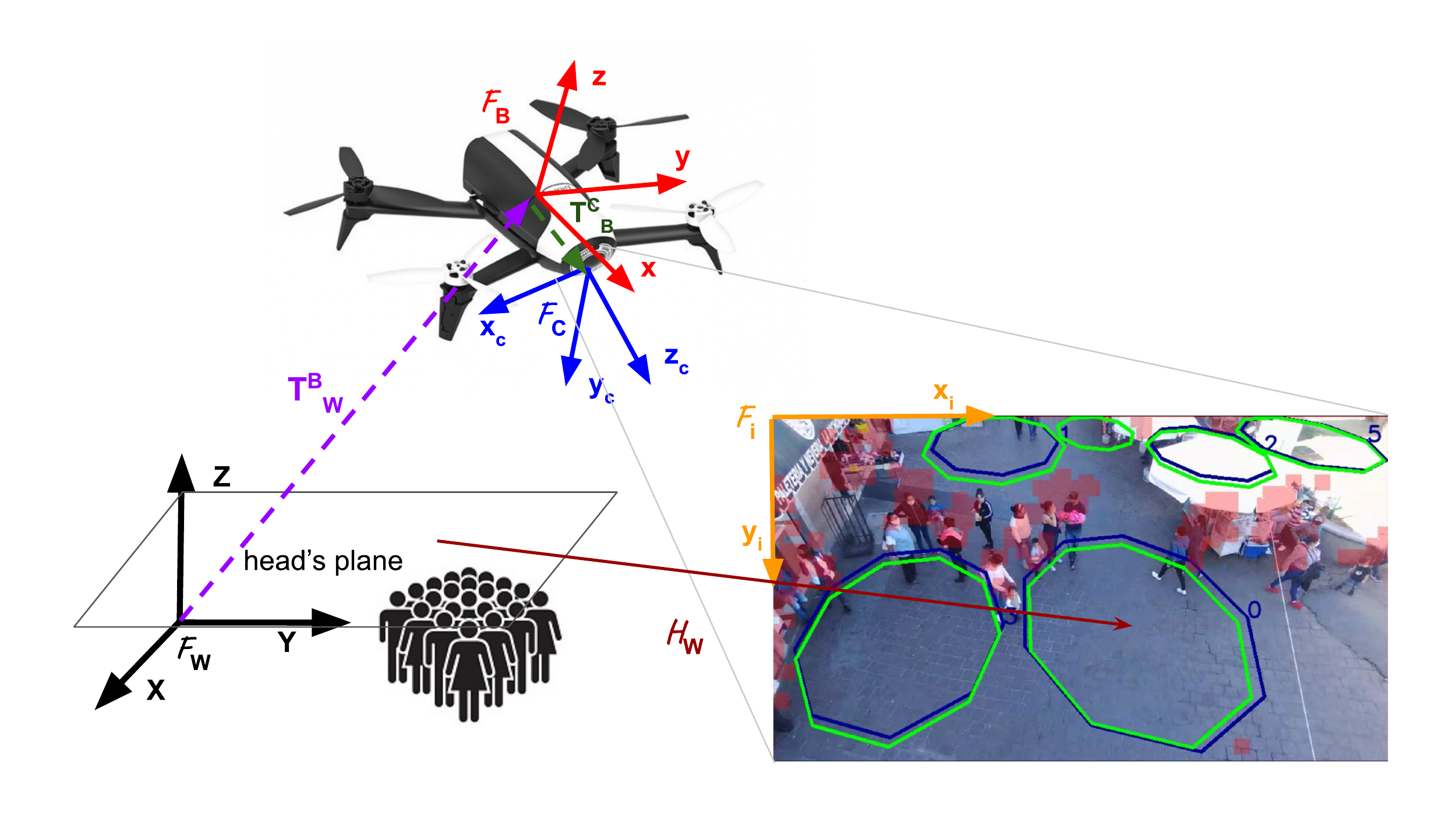}}
\caption{UAV moving in a populated environment. Three reference frames are defined, an inertial one fixed to the ground, a body fixed frame attached to the UAV and a camera frame. Then, the camera pose can be retrieved from the UAV on-board sensors measurements.}
\label{fig:reference_frame}
\end{figure}

The objective of our algorithm is to propose SLZ in populated environments to prevent the drone from hurting people in case of a system failure. For that matter, we consider a  camera attached to an UAV. Moreover, people may freely  move  in  the  scene  with  unknown  dynamics. For now, we restrict our study to scenes where people are primarily moving on a horizontal main plain.

In the present modeling, three reference frames are used. The first one is the inertial or global reference frame, which is denoted as $\mathbf{F_W}$. The body reference frame is attached to the UAV's center of mass and is referred as $\mathbf{F_B}$. The homogeneous transformation from the UAV's body reference frame $\mathbf{F_B}$ to the global reference frame $\mathbf{F_W}$ is denoted by $\mathbf{T}^W_B \in SE(3)$ and is assumed to be retrievable by means of the UAV's on-board sensors. The third reference frame uses the camera center as its origin and is denoted as $\mathbf{F_C}$, and its transformation with respect to the body fixed frame $\mathbf{T}^B_C \in SE(3)$ is known a priori. Fig.~\ref{fig:reference_frame} shows the three reference frames and their transformations.

An important element in the proposed solution is the so-called head plane, $P$, which is a plane in the scene parallel to the floor that crosses the average height of a person at $h_h$. More precisely, let $P$ be the plane defined in $\mathbf{F_W}$, with $P_0 = (0,0,h_h)$ and $\Vec{\bf{n}} = (0,0,1)$ defining its point-normal form. The head plane $P$ is crucial since computations of SLZ proposals are performed on it. Additionally, let $I$ be the image plane as customary defined in $\mathbf{F_C}$.

The algorithm is divided into two main stages: the SLZ candidates proposal and the multiple SLZ tracking that ensures time consistency between frames. The former is described in this section, while the latter in Section~\ref{sec:tracking}. In the SLZ proposal stage, we obtain at most $N_{p}$ SLZ candidates in the $P$ plane, with a radius of at least $r_0$ meters. Towards the generation of these SLZ proposals, we first calculate the density map $D$ in the image plane $I$ using a density map generator. For our experiments, we use the the lightweight deep learning architecture Pruned BL CCNN \cite{gonzalezLandingSite}.

The next step is to obtain an occupancy map $O$ in the $I$ plane, based on the density map $D$. The map $O$ has the same size as $D$, where each pixel has either a value of $255$ or $0$ denoting whether there is a person or not according to the density values from $D$. For that, all the occupancy map pixel values $O_{i, j}$ are set according to the next expression:
\begin{equation*}
	    O_{i, j}			=		
	    \begin{cases}
        255 & D_{i, j} - \min\{D_{i, j}\} > 0, \\
        0 & D_{i, j} - \min\{D_{i, j}\} = 0.
\end{cases}\\
    \end{equation*}
After that, the zones with values equal to $255$ are dilated two times using a kernel of ones with shape $\mathbb{R}^{5\times5}$ to further overestimate the crowd's location for security reasons. Finally, we perform a bitwise-not operation to set all values equal to $0$ to $255$ and vice versa.

Thereafter, the occupancy map $O$ is projected to the head's plane $P$ based on the rigid transformations that relate reference frames $\mathbf{F_W}$, $\mathbf{F_B}$ and $\mathbf{F_C}$, which are obtained from the UAV's sensors and the a priori known camera's pose with respect to the UAV's body. These assumptions are consistent with the information commonly available from most UAVs. More precisely, let $\mathbf{T}^B_W \in SE(3)$ be the rigid transformation from inertial frame coordinates to body frame coordinates, and $\mathbf{T}^C_B \in SE(3)$ the transform from body frame coordinates to camera frame coordinates (see Fig. \ref{fig:reference_frame}). Thus, denoting the intrinsic camera matrix by $\mathbf{K}$, the projection from a point $(x_W,y_W,h_h,1)^T$ in the $P$ plane to a point $(x_I,y_I,1)^T$ in the image plane $I$ is given as:
\begin{equation} \label{eq:proj}
\lambda \left[
   \begin{array}{c}
        x_I  \\
        y_I \\
        1   
\end{array}\right]
   =  \mathbf{K}  \mathbf{T}^C_B \mathbf{T}^B_W
  \left[
  \begin{array}{c}
        x_W  \\
        y_W \\
        h_h  \\
        1   
\end{array}\right]
, 
\end{equation}
with $\lambda$ a scale factor, which can be easily retrieved from the UAV altitude and the camera pose.

Additionally, to perform the projection of the occupancy map $O$ to the head plane $P$, denote it as $O^W$, a grid $G \in P$ is defined. Each element $G_{i,j}$ of the grid stores a coordinate $(x_W, y_W)$ on the $P$ plane; refer to them as $(G^{x}_{i,j},G^{y}_{i,j})$. Then, the corresponding occupancy values for $O^W$ are retrieved as follows. Utilizing Eq.~(\ref{eq:proj}) and given a grid element $G_{i,j}$, its stored inertial coordinates $(G^{x}_{i,j},G^{y}_{i,j})$ are mapped to an image coordinate $(x_I,y_I)$ as: 
\begin{equation}
   \lambda \cdot (x_I, y_I, 1)^T = \mathbf{K}  \mathbf{T}^C_B \mathbf{T}^B_W \cdot (G^{x}_{i,j},G^{y}_{i,j}, h_h, 1)^T. 
\end{equation}
The image coordinate $(x_I,y_I)$ is employed to query the occupancy map $O$. Thus, the respective occupancy value in $O$ is used as the occupancy value for $O^W_{i,j}$. The implementation of the past procedure is based on the sampler module of the Spatial Transformer Network \cite{jaderberg2015spatial}.

\begin{figure}[t]
    \centerline{\includegraphics[width=0.45\textwidth]{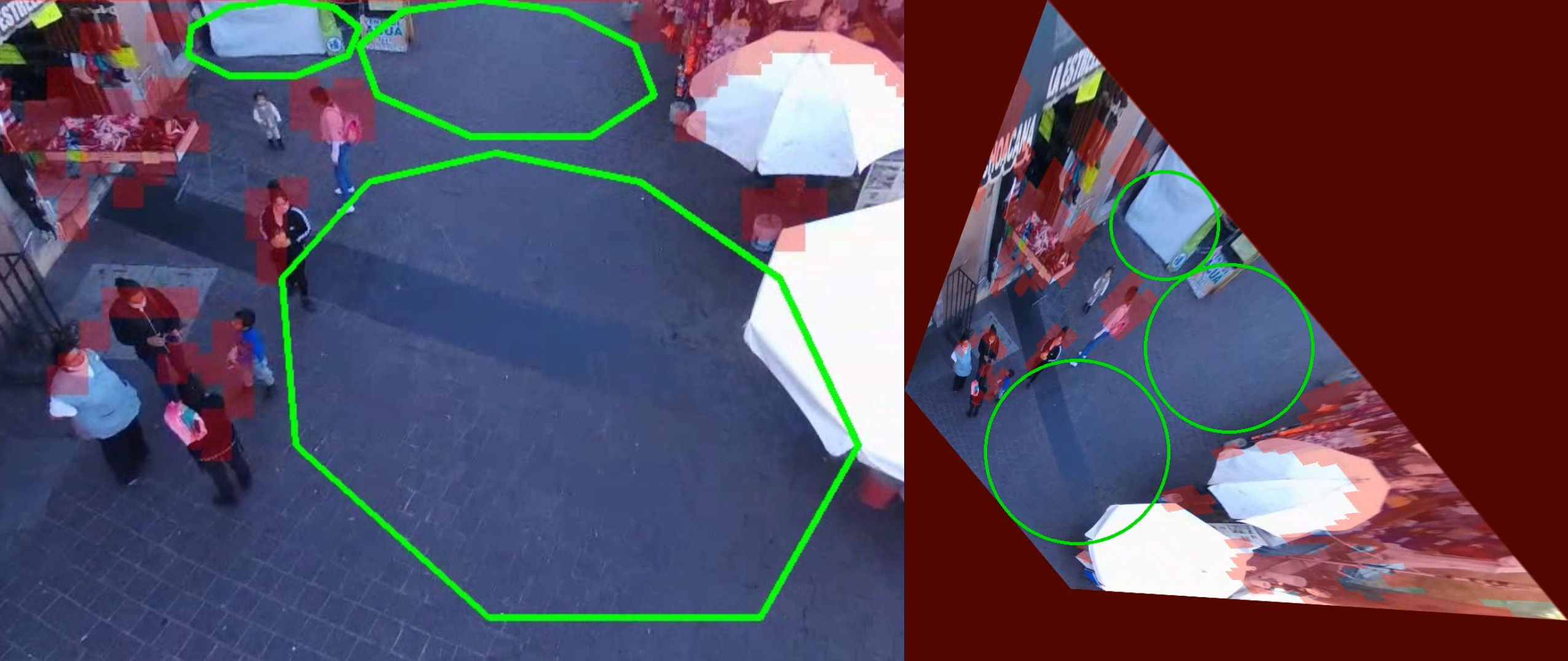}}
    \caption{Safe landing zones proposals in the image plane (left) and its corresponding projection in the head's plane (right). The safe landing zones are found in the head's plane using the projected occupation map.}
    \label{fig:projection}
\end{figure}

In the final step, we obtain a distance map $C$ from the occupancy map $O^W$ in the head's plane. The distance map $C$ considers the binary map encoded in $O^W$, and stores the distance of each pixel with non zero value (people free) to the the nearest zero value pixel (occupied by people). By finding the position and the value of the pixel, we effectively find the biggest circular landing zone center and its radius in the head plane $P$. Next, we mark the landing zone in the projected occupation map $O^W$ as occupied, and repeat the process until a number of $N_p$ of SLZ candidates are encountered, or until a small landing zones with radius less than $r_0$ is obtained. The results of the SLZ detection algorithm in the image plane $I$ and in the holography projection to the head's plane $P$ are illustrated in Fig. \ref{fig:projection}.

\section{Multiple landing zone Tracking}
\label{sec:tracking}

By themselves, the safe landing zone proposals could be used to indicate where a UAV could land. Nonetheless, due to the difficulty of the task where people in the scene move freely with unknown dynamics within the plane $P$, in addition to the density maps not being exempt to fail due to environmental variations, %
the SLZ candidates might critically change from a frame in time $k$ to a frame in time $k+1$. This could be catastrophic in the context of an emergency landing mission. Therefore, as the second stage of our algorithm, we propose to use multiple instance KF trackers to filter out outliers in the SLZ detection, smoothing out the movement of the landing regions, preventing abrupt jumps and ensuring temporal consistency along frames.

For a given instance $i$ of a SLZ encountered using the procedure described in the previous section, let us consider a discrete time KF, with state vector at time $k$ equal to
\begin{equation}
\mathbf{x}_{k}^{(i)} =  \left( x_k^{(i)},y_k^{(i)},r_k^{(i)},\Dot{x}_k^{(i)},\Dot{y}_k^{(i)}, \Dot{r}_k^{(i)} \right)^T,    \end{equation}
where $x_k^{(i)}$ and $y_k^{(i)}$ represent the coordinates of the SLZ circle's center in the head plane $P$, $r_k^{(i)}$ is the circle radius, while the velocity of the coordinates of the center and the rate of change of the radius are given by $\Dot{x}_k^{(i)}$, $\Dot{y}_k^{(i)}$ and $\Dot{r}_k^{(i)}$, respectively.

For the dynamic model, we use a constant velocity model with acceleration considered as the process noise $\mathbf{\omega}$ with normal distribution \cite{saho18_kalman_filter_movin_objec_track}, that is,
\begin{equation}
    \mathbf{x}_{k+1}^{(i)} = 
    \begin{bmatrix}
        \mathbf{I}_3 & \Delta t \mathbf{I}_3\\
        \mathbf{0}_3 & \mathbf{I}_3
    \end{bmatrix}
    \mathbf{x}_{k}^{(i)}+
 \mathbf{\omega}\ \ \ |\ \ \ \mathbf{\omega}\sim \mathcal{N}(\mathbf{0},\,\mathbf{Q})    ,
\end{equation}
with $\mathbf{0}_3\in \mathbb{R}^{3 \times 3}$ a matrix of zeros, $\mathbf{I}_3 \in \mathbb{R}^{3 \times 3}$ the identity matrix, and $\Delta t$ the time increment. The process noise covariance matrix $\mathbf{Q}$ is obtained as:
\begin{equation}
 \mathbf{Q}=\sigma_a \begin{bmatrix}
 \frac{\Delta t^4}{4} \mathbf{I}_3& \frac{\Delta t^3}{2} \mathbf{I}_3\\
\frac{\Delta t^3}{2} \mathbf{I}_3& \Delta t^2 \mathbf{I}_3
\end{bmatrix} ,
\end{equation}
where $\sigma _a$ is the acceleration uncertainty and is selected empirically.

The measurement model considers the observation vector $\mathbf{z}^{(i)}_k$ as the coordinates of the center of the SLZ and its radius, as obtained by the algorithm described in Section \ref{sec:proposals}, with a measurement noise $\mathbf{\nu}$ with normal distribution, therefore, the measurement model is given by:
\begin{equation}
    \mathbf{z}^{(i)}_k = 
    \begin{bmatrix}
        \mathbf{I}_3 & \mathbf{0}_3
    \end{bmatrix}
    \mathbf{x}^{(i)}_k + \mathbf{\nu}\ \ \ |\ \ \ \mathbf{\nu}\sim \mathcal{N}(\mathbf{0},\,\mathbf{R})    ,
\end{equation}
with $\mathbf{R}$ standing for the measurement covariance matrix.
To handle multiple target tracking and associate the main SLZ candidates with their corresponding trackers, the Hungarian algorithm is employed, a combinatorial optimization algorithm that solves the assignment problem in polynomial time \cite{bruff2005assignment}.
We associate a SFZ proposal with a KF by creating a cost matrix using the Intersection over Union ($IoU$) criteria, between all the combinations of the SFZ measurements and the KFs estimates. For two circle areas $A_i$ and $A_j$, the IoU is computed as
\begin{equation}
    IoU = \frac{A_i \cap A_j}{A_i \cup A_j}.
\end{equation}
Then, the cost matrix is fed to the Hungarian algorithm, which associates the ``most similar" SFZ proposal for each KF tracker. If no match is found for a KF, it is assumed that no consistent observation is available for that filter, and a no-found counter is increased by one. When the counter is greater than a threshold $\mu_1$, the respective KF is eliminated. If an unmatched SFZ proposals is found for a number of $\mu_2$ consecutive iterations, and the number of KFs is below $N_p$, a new KF is initialized.

At the beginning of a trial, a maximum of $N_p$ KFs are created by initializing their states using the first SFZ proposals. In the following frames, the propagation step is performed, and when available, the predicted state is corrected through the corresponding observation.

\begin{figure*}[h!]
    \centering
    \begin{tabular}{ccc}
        \subcaptionbox{\label{a}}{\includegraphics[width = 2.2in]{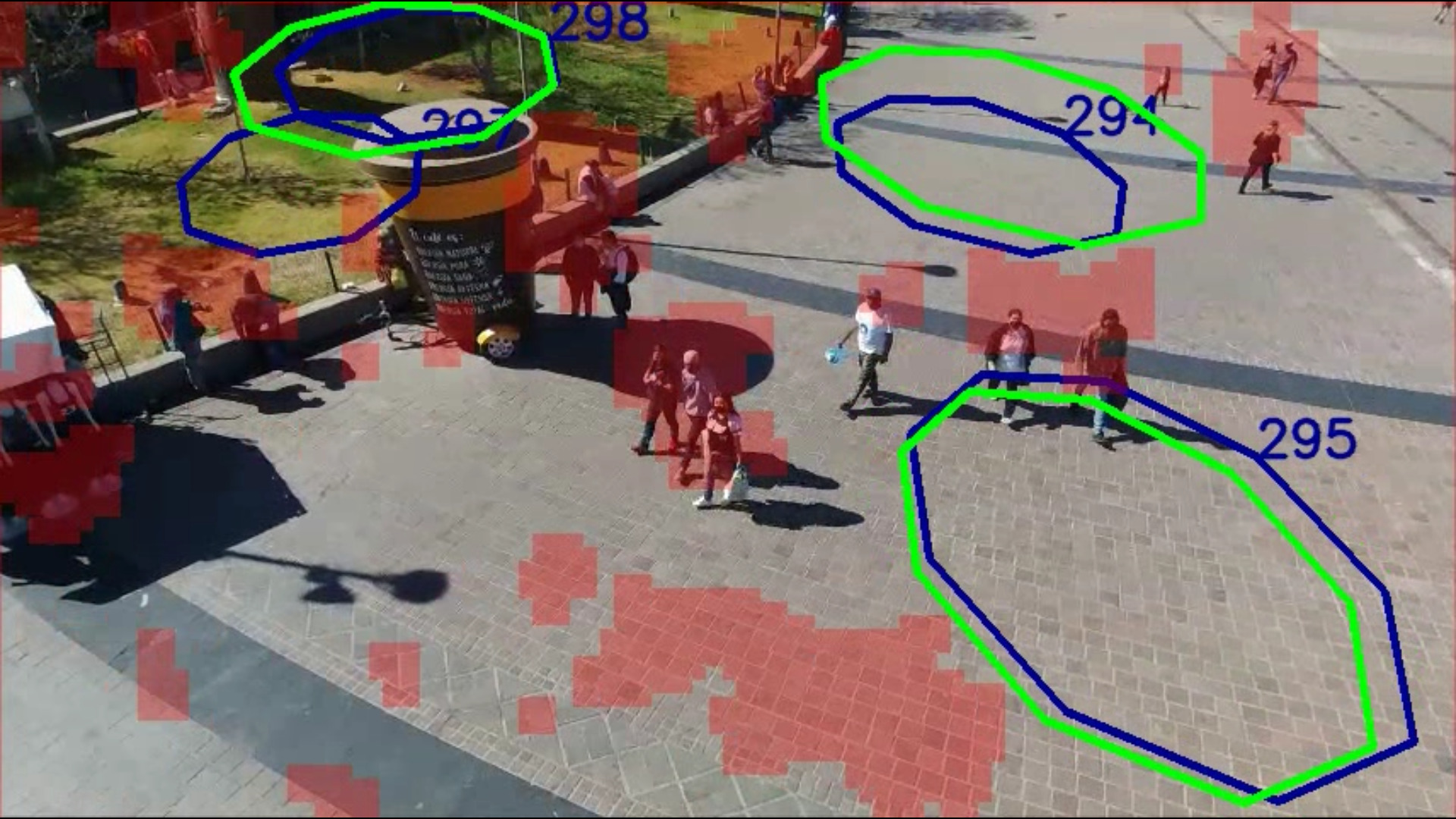}} &
        \subcaptionbox{\label{b}}{\includegraphics[width = 2.2in]{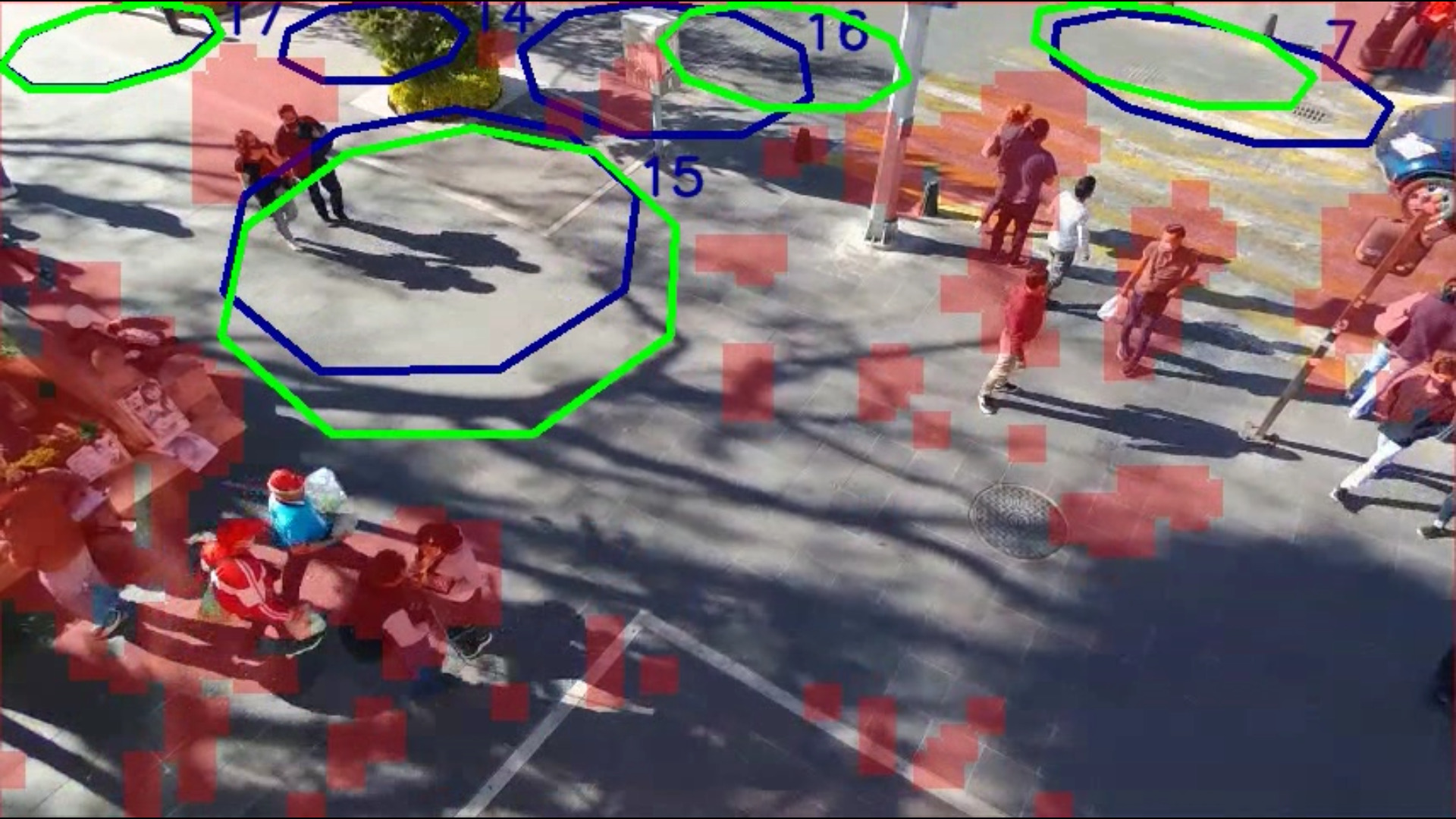}} &
        \subcaptionbox{\label{c}}{\includegraphics[width = 2.2in]{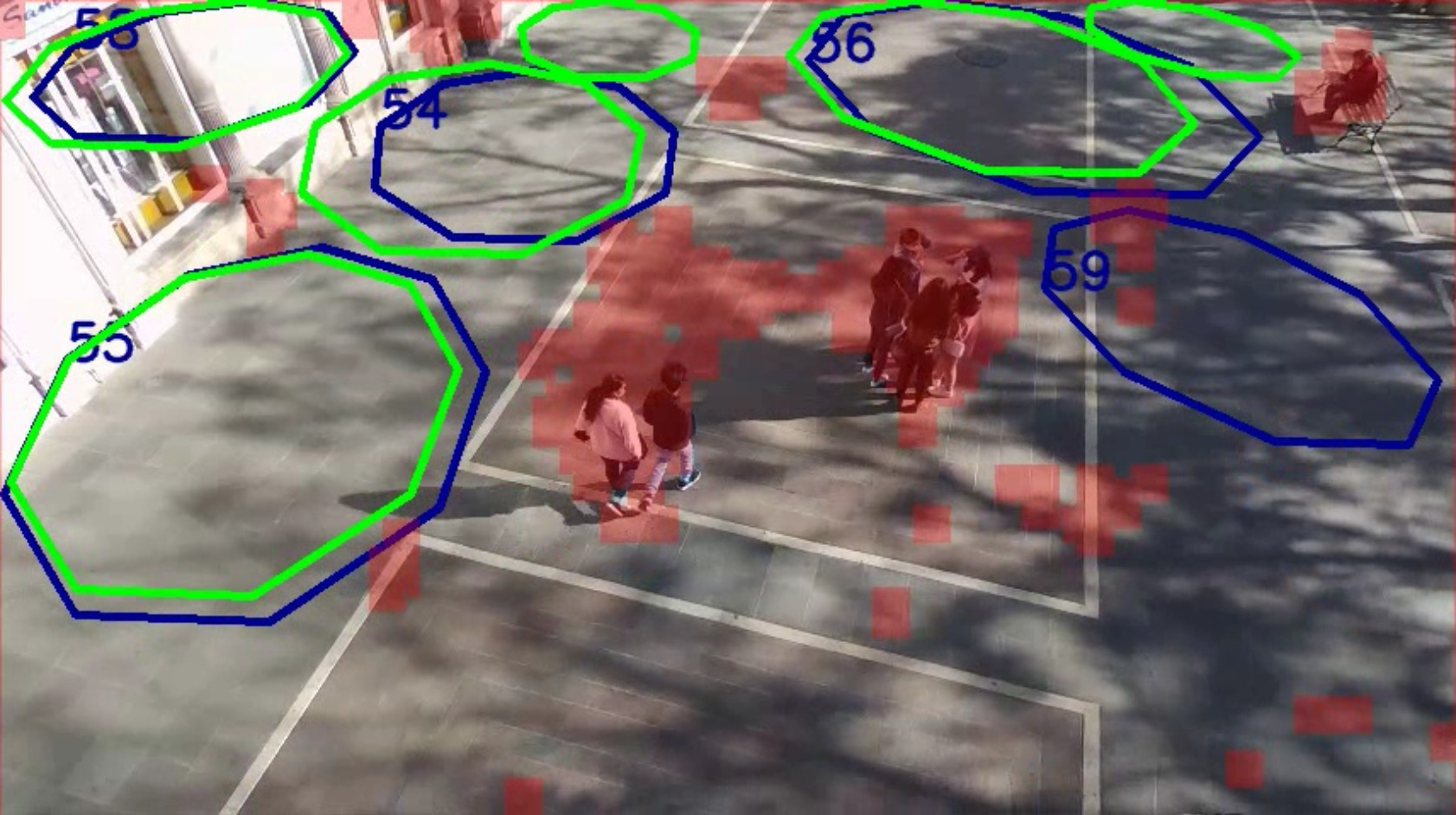}} \\
   \vspace{-0.3cm}     
        \subcaptionbox{\label{d}}{\includegraphics[width = 2.2in]{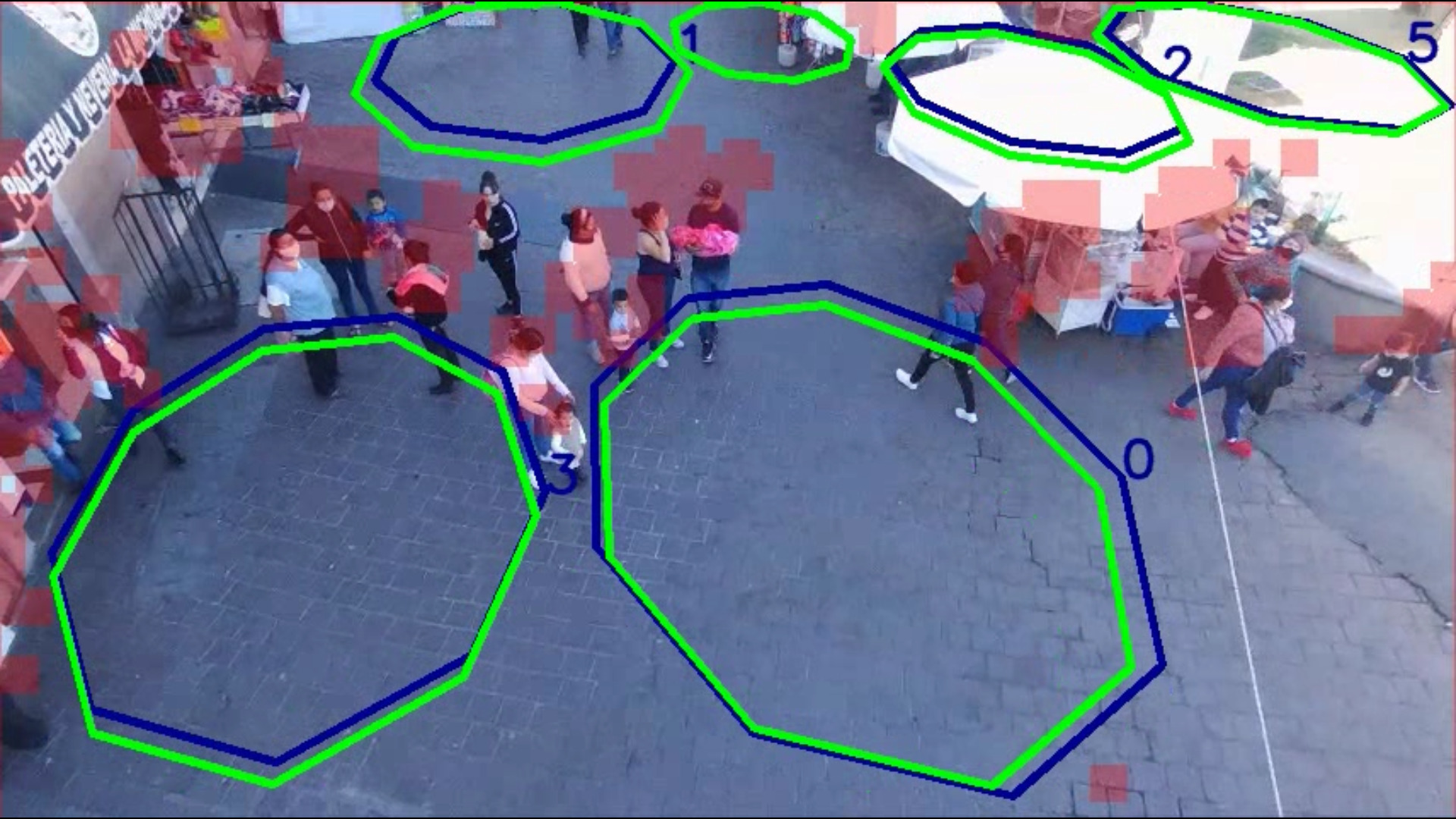}}  &
        \subcaptionbox{\label{e}}{\includegraphics[width = 2.2in]{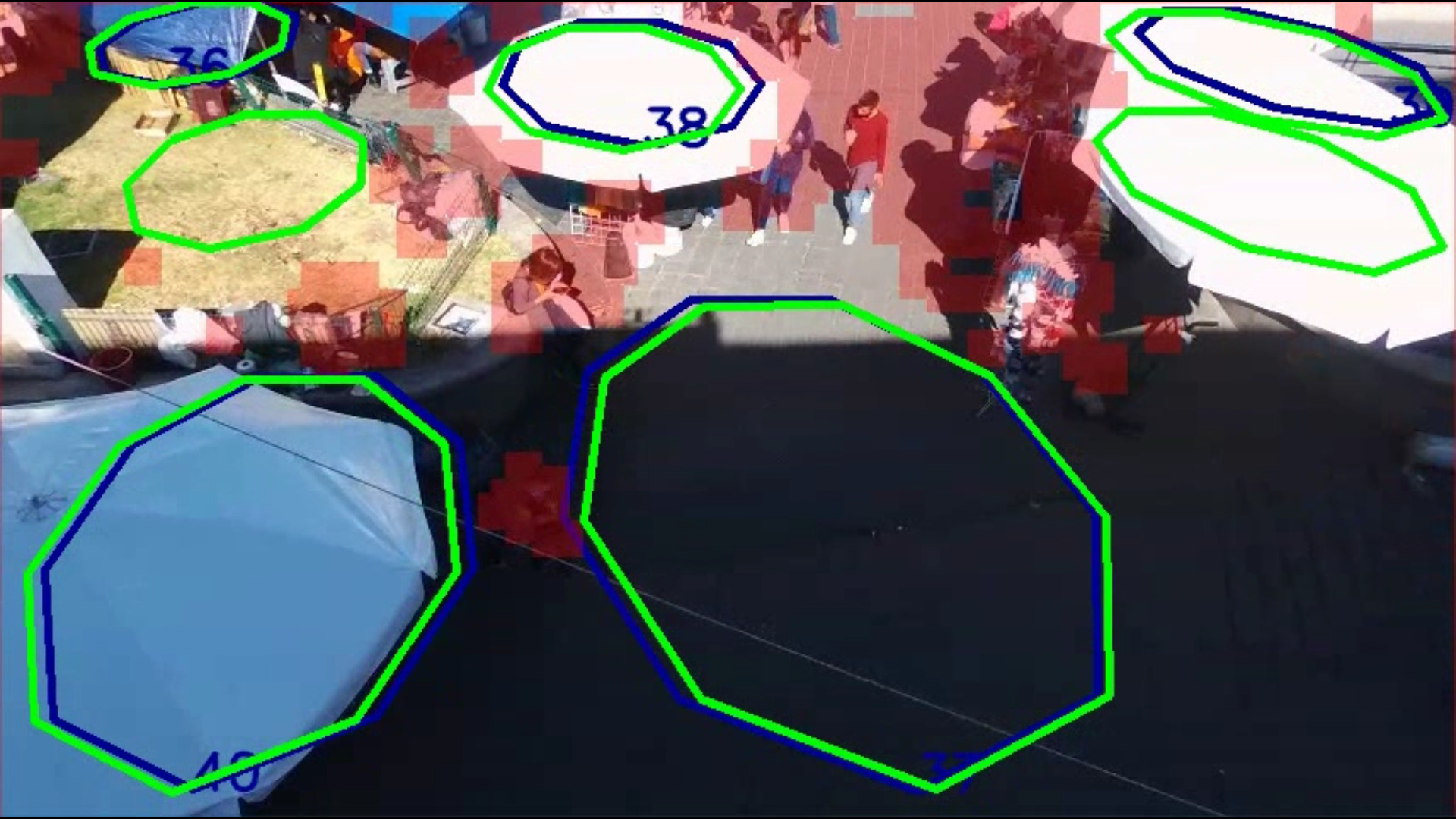}} &
        
        \subcaptionbox{\label{f}}{\includegraphics[width = 2.2in]{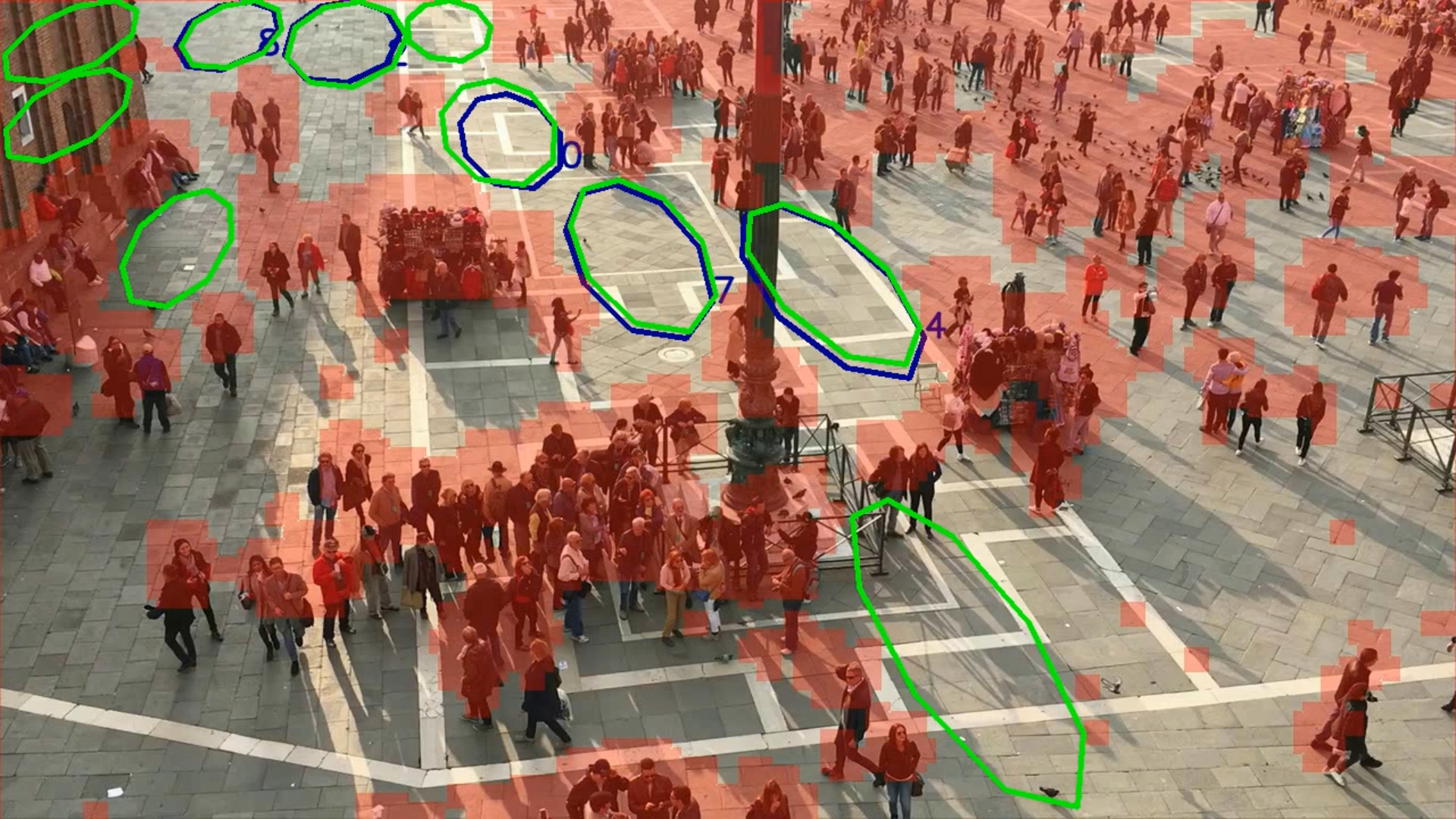}}\\
        
    \end{tabular}
    \caption{Experiments results. The red zones are where the crowd is located, the green circles are the current frame safe landing proposals and the blue circles are the trackers. As shown in the images, the goal is to find suitable landing spots for the drone to land in a variety of scenarios. From sparse crowds (Fig \ref{a}) to low illumination near the ground (\ref{e}).}
    \label{fig:experiments_results}
\end{figure*}

\section{Experimental Validation}
\label{sec:experiments}

The proposed strategy was validated in a wide variety of real  scenarios recorded from a drone in public squares, as well as with videos from the Venice dataset \cite{Liu_2019_CVPR}. The former scenarios where recorded by our lab members using a Parrot Bebop 2. The drone  was remotely controlled using  ROS (Robotic Operating System) from a ground computer, from where videos and sensor data were recorded. The video resolution is set to $480$p at $15$ to $30$ fps, depending on the communication quality. We used the IMU (Inertial Measurement Unit), GPS (Global Positioning System) and altimeter filtered data available at $5$ Hz to obtain the drone's pose and use it to project the image to inertial frame coordinates. The camera is mounted in the front of the drone with a wide-angle lens. Additionally, the camera allows to select a region of interest with distortion correction by software using virtual tilt and pan angles. 

As shown in Fig. \ref{fig:experiments_results}, the scenes come from different points of view of two distinct public plazas taken from an UAV. These recordings display a variety of characteristics that make them specially challenging to test safe landing zone proposal algorithms. All of them contain moving crowds with changes in illumination, heterogeneous surfaces to land, different drone's altitude and movement, and different backgrounds.

In the dataset experiments, we used three scenarios from the Venice dataset \cite{Liu_2019_CVPR} recorded in the San Marcos plaza. These videos were recorded using a cellphone above the crowd at a resolution of $720$p at $25$ fps. Since no camera pose was obtained, the homographies were estimated from the image alone at every $60$ frames mark. Since the homography has a update rate of 0.5 Hertz, we fix in place the homography for short periods of time where it did not appear to be significant camera movements. 
All of the scenes present a dense crowd moving in a flat surface which make them suitable to test the SLZ proposal algorithm despite being created for the crowd counting task.

Fig. \ref{fig:experiments_results} also shows the performance of our algorithm in different scenes. We see in green the current frame SLZ candidates as computed by the algorithm described in Section \ref{sec:proposals} and in blue the KF trackers estimate. In all figures, it can be seen that the algorithm, for that frame, locates suitable SLZ proposals and trackers in zones free of people. The challenge that each scene offers varies depending on the characteristics described above. For example, in Figure \ref{e} we see one of the most challenging scenarios; where the camera is near to the ground in a narrow passage, capturing people in constant flow, making it difficult to keep track of the landing zone proposals, while presenting high illumination contrasts due to the sun. Despite of this, the oldest trackers (signaled by a lower number) keep track of umbrellas and a small green area where the drone could land without hurting people.
Also, in Fig. \ref{f}, we display an scene from the Venice dataset, showing good performance since all of the trackers cover a valid SLZ proposal free of people.

Another interesting aspect can be observed in Figs. \ref{a}, \ref{b}, \ref{c}, where sparser crowds are presented in a wide scene. We can observe how the rich textures on the floor may confuse the density map appearing as false positive crowds. This is produced since the density map generators are explicitly trained to overestimate the crowds to increase the algorithm security. Nevertheless, the proposed strategy was able to find suitable SLZ in the scenes. A more robust density map generator, such as DM-Count \cite{wang20_distr_match_crowd_count}, could help to mitigate this phenomena at a computational cost, compromising also the capacity to implement them in real-time embedded applications.

Finally, in order to better appreciate the performance of the strategy in dynamic scenarios, we provide a video in the multimedia material of this paper, containing the results of our algorithm in the various scenarios here presented.

\section{Conclusions and Future Work}\label{sec:conclusions}%

In this paper we present a novel algorithm for visual-based autonomous safe landing of UAVs in populated environments. We believe that this kind of algorithms are a key aspect for the successful application of UAVs in urban areas, provided that they will help to prevent injuring people in case of emergency landing.

The proposed algorithm, in conjunction with an adequate density map generator, is capable of finding Safe Landing Zones proposals in a variety of real uncontrolled and challenging scenes where the crowds move with unknown dynamics, with illumination changes, different crowd densities and rich image features. Moreover, with the help of the KFs, the algorithm is able to keep track of Safe Landing Zones proposals even when the density map fails to correctly detect parts of the crowds in consecutive frames. 

Although the present work showed promising results, it is not exempt from failure, and due to the hard security constrains involved and the wide diversity of challenging situations that may occur during and emergency landing mission, further work is still required to obtain a fully reliable solution. Future efforts, will be devoted to define a suitable criteria to choose the best SLZ available. Also, it is required to impose a horizontality condition on the main planes of the scene in order to discriminate unsuited regions for landing, such as walls or steep terrains. Another interesting approach to further increase the people's security would be to add a human body detector as the UAV comes closer to the ground. Also, at present we focus all the efforts in avoiding to hurt people, but preventing the UAV from crashing with other obstacles would be also of interest. Finally, the real time capabilities could be further improved by having an end-to-end deep learning solution by training it using synthetic and real data.

\bibliographystyle{IEEEtran}
\bibliography{references}

\end{document}